%% file: manuscript.tex
\documentclass[11pt]{article}
\usepackage[T1]{fontenc}
\usepackage[utf8]{inputenc}
\usepackage{lmodern}
\usepackage[margin=1in]{geometry}
\usepackage{graphicx,booktabs,longtable,array,tabularx,amsmath,amssymb}
\usepackage[numbers,sort&compress]{natbib}
\usepackage{xurl}
\usepackage[hidelinks]{hyperref}
\usepackage{microtype,enumitem}
\setlist{nosep,leftmargin=*}
\newcolumntype{P}[1]{>{\raggedright\arraybackslash}p{#1}}
\newcolumntype{Y}{>{\raggedright\arraybackslash}X}
\newcommand{\code}[1]{\texttt{\detokenize{#1}}}
\title{DR-LabStack: Design and Implementation of a Clinician-Facing Web System for Diabetic Retinopathy Prediction}
\author{\normalsize Yingfan Xu$^{1}$\thanks{Corresponding author.}\quad Tieming Liu$^{1}$\quad Ye Liang$^{2}$\\[0.5em]
\small $^{1}$School of Industrial Engineering and Management, Oklahoma State University\\
\small $^{2}$Department of Statistics, Oklahoma State University}
\hypersetup{pdftitle={DR-LabStack: Design and Implementation of a Clinician-Facing Web System for Diabetic Retinopathy Prediction},pdfauthor={Yingfan Xu, Tieming Liu, Ye Liang}}
\date{}
\begin{document}
\maketitle
\begin{abstract}
Pretrained diabetic retinopathy (DR) prediction models differ in their input fields, serialization formats, preprocessing requirements, and output semantics. Making these models accessible through a common clinical interface therefore requires explicit coordination between the user interface and the inference service. We designed and implemented DR-LabStack, a React--Flask web system integrating four externally developed pretrained models: RuleFit, Pruned RuleFit, Elaborative XGBoost, and Two-level Ensemble. A shared form retrieves ordered model features, renders model-specific numerical and categorical controls, and constructs a positional input vector. Backend adapters load heterogeneous artifacts and apply the ensemble's accompanying scaler, while a common JSON response supports binary classification display alongside method and source information. Functional evaluation on September 8, 2026 used copied application files and real model artifacts in a documented isolated environment. All four models loaded and exposed their 14-, 6-, 8-, and 25-field contracts. Sixty-two Flask test-client requests characterized service behavior; 12 limited-vector checks confirmed invocation-path and threshold consistency. Twenty-four browser-component scenarios with mocked transport verified input ordering and result rendering and characterized input-validation behavior. The resulting system demonstrates a reusable interaction and serving workflow for heterogeneous DR models. The contribution is web-system design, integration, and software functionality; clinical effectiveness and clinician usability require separate evaluation.
\end{abstract}

\section{Introduction}
Diabetic retinopathy (DR) research includes image-based grading methods such as HAPM, which uses pathological descriptions to refine visual prototypes across medical centers \citep{zhu2025pathology}. Prediction from structured clinical data has motivated research using laboratory measurements, recorded complications, and demographic variables. Related studies have also developed web tools that make predictive models accessible through manual data entry \citep{gandhi2023,wan2025,he2026}. For a research group working with several such models, the software task extends beyond exposing a single prediction function. Each model can require a different ordered feature vector, artifact loader, preprocessing object, and interpretation of its numerical output. A common interface must preserve these distinctions while presenting an understandable interaction. Model reuse and deployment are also central concerns in broader software-engineering studies of machine-learning applications \citep{amershi2019se}.

We designed and implemented DR-LabStack to support this workflow across four pretrained models: RuleFit, Pruned RuleFit, Elaborative XGBoost, and Two-level Ensemble. The application combines React model pages and a shared prediction form with Flask model-discovery, feature-discovery, and inference endpoints. The project repository is publicly available.\footnote{\url{https://github.com/TerrificXu/diabetic-retinopathy-web}} This paper describes the complete four-model system and evaluates its implemented integration paths.

The predictive models were developed by their respective researchers in separate prior work. The contribution of this paper is the design, implementation, and functional evaluation of the web application and its integration interfaces. Algorithm design, training, feature selection, rule pruning, and ensemble learning belong to the underlying model research. We neither introduce a new prediction algorithm nor retrain the supplied artifacts in this study.

Three engineering concerns motivate the design. First, positional inference APIs make feature order a correctness condition: recognizable labels alone cannot ensure that a value reaches the expected model input. Second, heterogeneous artifacts require model-specific loading and preprocessing, including an external scaler for the ensemble. Third, a uniform display must distinguish a shared response format from a shared probabilistic meaning. These concerns connect interface design directly to the serving layer.

The system contributions are: (i) a shared React interaction that uses model-derived ordering while retaining model-specific display labels, hints, and categorical controls; (ii) a Flask integration layer for heterogeneous pretrained artifacts, associated preprocessing, and a common classification response; and (iii) a functional evaluation separating real-model service checks from browser-component tests. The evaluation establishes implemented behavior for specified inputs and identifies the integration contracts needed for further use. The interface is clinician-facing in its intended audience and presentation; clinician participation and clinical utility were not evaluated in this study.

\section{Related Work}
\subsection{Structured-data DR prediction and web delivery}
Gandhi, Daskivich, and Ogunyemi describe DRRisk, a Flask-based tool for assessing current DR risk using an EHR-derived model \citep{gandhi2023}. Their study includes clinical consultation and presents probability-based risk categories. This work provides directly relevant precedent for delivering non-image DR predictions through a web interface. DR-LabStack focuses on the integration of multiple model-dependent input and artifact contracts within a common interaction.

Wan et al. study DR prediction using routine laboratory tests and examine an XGBoost model with SHAP \citep{wan2025}. He et al. report a multicenter laboratory-based prediction study and a Streamlit application with four inputs, shared preprocessing, probability output, and individual SHAP visualization \citep{he2026}. These studies address model performance in defined cohorts and, in the latter case, a model-specific deployment interface. Our evaluation concerns software integration across four supplied models. Original-study diagnostic estimates are not pooled, ranked across cohorts, or presented as reproduced results of this system.

\subsection{Underlying models and interpretability}
RuleFit represents predictions through an additive combination of rule indicators and linear terms \citep{friedman2008}. The pruning-and-merging research linked from the Pruned RuleFit page is attributed to Bani Ahmad and Liu \citep{baniahmad2025}. Laoh and Liu describe incorporating medical domain knowledge into DR model development \citep{laoh2024}; that research provides context for the eight-input Elaborative XGBoost artifact. XGBoost itself is an established gradient-boosting method \citep{chen2016}. Mahmoudi and Liu's two-level ensemble work develops nested stacking configurations for DR prediction \citep{mahmoudi2025}. DR-LabStack supplies the interaction and serving interfaces for these externally developed artifacts.

Model structure, predictor selection, methodological context, and individual explanations serve different purposes. The distinction between interpretable models and explanations of black-box predictions is discussed by Rudin \citep{rudin2019}. We use this distinction to describe what the integration exposes: a model may have inspectable rules or a domain-informed input subset while the application presents only field labels, method context, and a classification.

\subsection{Model serving and systems engineering}
General serving systems such as Clipper use modular interfaces to connect applications to heterogeneous machine-learning frameworks \citep{crankshaw2017}. DR-LabStack applies a small in-process architecture to a clinical research application, emphasizing ordered inputs, artifact adapters, and user-facing output semantics. Its implementation and evaluation scope differ from serving-infrastructure benchmarks concerned with throughput, caching, and distributed execution.

Amershi et al. describe an iterative development workflow spanning model requirements, data preparation, evaluation, deployment, and monitoring \citep{amershi2019se}. DR-LabStack addresses the application-integration part of that broader workflow, accepting pretrained artifacts and connecting them to a shared interface. Sculley et al. identify data dependencies and package-specific integration code as sources of maintenance difficulty in ML systems \citep{sculley2015}. In our application, ordered features, display definitions, and preprocessing objects make these dependencies concrete at the boundary between manual entry and model invocation.

\subsection{Clinical interfaces and human--AI interaction}
Amershi et al.'s interaction guidelines distinguish communicating system capabilities from communicating expected reliability \citep{amershi2019hai}. This distinction separates an operable prediction form from an adequate account of what its output means. In a qualitative study with 21 pathologists using a prostate-cancer AI assistant, Cai et al. identified needs for global information about model limitations and design objectives, beyond explanations of individual decisions \citep{cai2019}. That study concerns clinical onboarding in image-based pathology; DR-LabStack provides method and source context beside structured-data entry forms. Its current information panels offer a place to communicate model context, while their adequacy for clinicians remains an empirical question. The CHIL 2025 research-roundtable report likewise emphasizes tailoring explanations to stakeholder needs and fitting them into clinical workflows \citep{alsentzer2025chil}.

TRIPOD+AI guides reporting of prediction-model development and evaluation \citep{collins2024}, while DECIDE-AI addresses early live clinical evaluation of AI decision support \citep{vasey2022}. These frameworks inform the evidence needed for later diagnostic or clinical-workflow studies. The present evaluation is a software functional study with explicitly defined fixtures and interface boundaries.

\section{Intended Use and System Overview}
DR-LabStack provides a manual-entry workflow for selecting a DR model, supplying its required inputs, and viewing a binary classification. Inputs span laboratory measurements, recorded complications, age, and demographic fields. The interface is intended to make research models accessible to clinical personnel for demonstration and assessment. The present evidence does not establish a basis for patient-care decisions, replacing retinal assessment, or excluding DR after a low output. A prediction horizon or early-stage disease target has not been verified for the integrated artifacts.

The implementation supports four interaction requirements: discovering the selected model's ordered fields; presenting recognizable labels and appropriate numerical or categorical controls; applying the supplied inference and preprocessing path; and returning a consistent result display with method and source context. The system includes four configured model pages, a model-list page, Home, Members, and Contact. The common form is the main connection between page-level presentation and model-specific inference.

\section{System Architecture and Implementation}
\subsection{Component responsibilities}
\begin{figure}[t]
\centering\includegraphics[width=\linewidth]{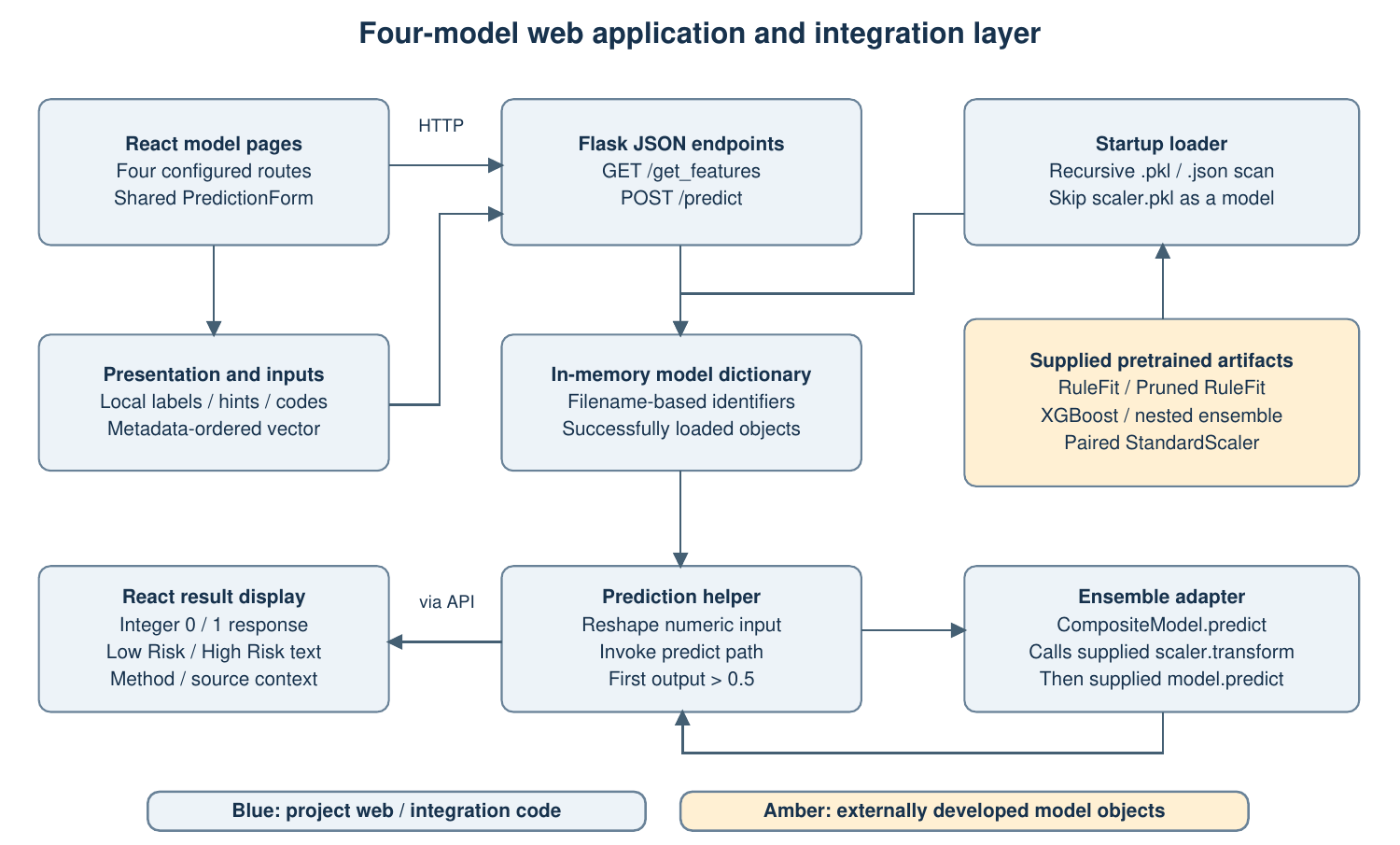}
\caption{DR-LabStack architecture for the complete four-model system. Blue boxes denote the web application and integration code developed in this project; amber boxes denote externally developed pretrained models and the paired scaler. The scaler path is used for Two-level Ensemble. Arrows describe implemented metadata and inference relationships; model training is outside this online workflow.}
\label{fig:architecture}
\end{figure}

Figure~\ref{fig:architecture} distinguishes application components implemented in this project from the pretrained artifacts and preprocessing object supplied by the model researchers. React handles routing, entry controls, input state, requests, and result rendering. Flask maintains the in-memory model dictionary, exposes feature metadata, and invokes an artifact-specific prediction path. Method descriptions, research links, and researcher information accompany the form as interface content.

The React entry point renders the application under StrictMode. Four model routes and their navigation links are explicitly configured, and each model page passes a fixed identifier to the shared \code{PredictionForm}. The model-list page separately requests the successfully loaded backend names. This arrangement supports shared form behavior while allowing model-specific page text. Adding an artifact can expand backend discovery, but a complete additional website page still requires route, navigation, and presentation configuration.

\subsection{Feature discovery and shared form construction}
When the selected model changes, the shared form requests \code{/get_features?model_name=...}. The response contains an ordered \code{features} array. The component uses exact names as state keys, generates controls by iterating over the array, and constructs the request vector in the same order. Display normalization is separate: frontend mappings provide readable labels, units, placeholders, and aliases for short or punctuation-heavy artifact names.

Numerical fields use required number inputs. Complication, gender, and race fields use required selectors with explicit numerical option values. Input hints remain frontend-maintained rather than supplied by a typed backend schema. Keeping presentation separate from exact artifact keys allows the same component to render different model contracts, while preserving responsibility for the meaning of units and categorical codes at the integration boundary.

\subsection{Artifact loading and preprocessing adaptation}
At startup, Flask recursively scans the relative \code{models/} directory for \code{.pkl} and \code{.json} files, skipping \code{scaler.pkl} as an independent model. A successfully loaded artifact is stored under its filename stem. Loading failures are reported and the corresponding entry is omitted. This mechanism is startup discovery; the implemented service has no hot-loading or online-upload operation.

The JSON adapter loads an \code{XGBClassifier} and obtains feature names from its booster. The Python serialization adapter tries joblib and then pickle. The helper contains additional format branches, but application discovery is limited to the two extensions above. Models remain in process memory for subsequent requests. The dictionary is an inference registry keyed by names, rather than a version-governance service.

An ensemble filename beginning with the normalized phrase ``two level ensemble'' triggers a lookup for a same-directory scaler. \code{CompositeModel} combines the predictor and scaler behind a common interface. It derives feature order from the model where available, otherwise from the scaler's \code{feature_names_in_}, and calls \code{scaler.transform} before prediction. The supplied ensemble uses this second metadata path. If the scaler is missing, the loader warns and returns the unwrapped model; the missing-scaler behavior is characterized in Section~\ref{sec:evaluation}.

\subsection{Request and response flow}
\begin{figure}[t]
\centering\includegraphics[width=\linewidth]{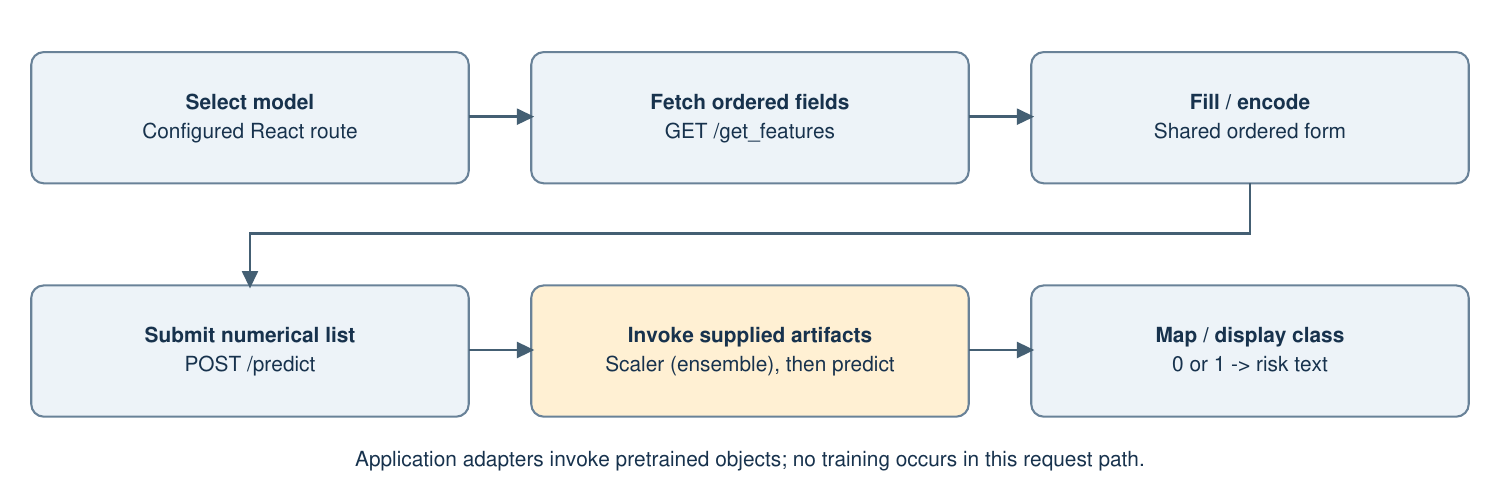}
\caption{Shared request sequence across all four models. Blue steps are implemented by the application. The amber step invokes the supplied predictor and, for the ensemble, its supplied scaler through the application adapter. The same metadata order drives rendered fields and the positional request.}
\label{fig:request}
\end{figure}

Figure~\ref{fig:request} follows the interaction from selection to display. For model $m$, let $F_m=(f_1,\ldots,f_{d_m})$ be the returned feature sequence and $u[f]$ the component's stored input string. The client sends
\begin{equation}
 x_m=(g(u[f_1]),\ldots,g(u[f_{d_m}])),\qquad
 g(v)=\text{\code{parseFloat(v) || 0}}.
\end{equation}
This expression describes JavaScript conversion, including a zero fallback for empty or unparseable values if conversion is reached. Native required controls govern ordinary blank-form submission; the fallback is not a model-derived missing-data procedure.

The \code{/predict} endpoint accepts a JSON model identifier and feature list. The helper converts the list to a floating-point array with shape $(1,-1)$, invokes \code{predict}, takes the first returned value, and applies a strict threshold:
\begin{equation}
 \widehat c_m=\mathbf{1}\!\left[\operatorname{first}\{\operatorname{predict}_m(P_m(x_m))\}>0.5\right],
 \label{eq:output}
\end{equation}
where $P_m$ is the external scaler for the ensemble and the identity operation for the other application adapters. Internal estimator transformations are retained within the supplied model. The JSON response contains an integer \code{prediction}; React maps 0 and 1 to ``Low Risk of DR!'' and ``High Risk of DR!'' respectively. This common response structure simplifies rendering while preserving the need to explain the underlying output semantics.

\begin{table}[t]
\small\centering
\caption{Main integration endpoints. Feature order is returned as metadata; field meanings and option labels are maintained by the frontend.}
\label{tab:api}
\begin{tabularx}{\linewidth}{P{.24\linewidth}P{.31\linewidth}Y}
\toprule Endpoint & Request & Response / behavior\\\midrule
GET \code{/models} & No payload & Names of successful startup loads.\\
GET \code{/get_features} & Query \code{model_name} & Ordered \code{features} array; unknown model returns an empty array with 200.\\
POST \code{/predict} & JSON \code{model} and \code{features} list & Integer \code{prediction}; invalid envelopes return 400; prediction failures return 500.\\\bottomrule
\end{tabularx}
\end{table}

Auxiliary pages organize program, researcher, and contact information. Separate Flask-WTF template paths also coexist with the React workflow. These supporting paths do not alter the shared React inference contract; their implementation details and observed ancillary failures are summarized in Appendix~\ref{app:ancillary}.

\section{Integrated Models and Interface Semantics}
\subsection{Model integration contracts}
Table~\ref{tab:models} lists all four integrated pretrained models and their source context. Field counts refer to inputs rather than laboratory tests. Appendix~\ref{app:fields} provides complete numbered keys, display meanings, and units/options as part of the manuscript's numerical contract.

\begin{table}[t]
\centering\small
\caption{Integrated pretrained models. L: laboratory measurements; C: complication fields. Source citations attribute the underlying methods, not an independently reproduced diagnostic result or a verified experiment identity for each artifact.}
\label{tab:models}
\begin{tabularx}{\linewidth}{P{.20\linewidth}P{.23\linewidth}P{.21\linewidth}Y}
\toprule Model / source context & Inputs & Artifact / external preprocessing & Output used by the helper\\\midrule
RuleFit \citep{friedman2008} &14: 12 L + 2 C & Pickle; no external scaler & Regression-mode \code{predict} score.\\
Pruned RuleFit \citep{baniahmad2025} &6: 4 L + 2 C & Pickle; no external scaler & Regression-mode \code{predict} score.\\
Elaborative XGBoost \citep{laoh2024,chen2016} &8: 5 L + age + 2 C & JSON; no external scaler & Classifier \code{predict} label.\\
Two-level Ensemble \citep{mahmoudi2025} &25: 20 L + age + 2 C + gender + race & Pickle and paired StandardScaler & Nested classifier \code{predict} label.\\\bottomrule
\end{tabularx}
\end{table}

Both RuleFit artifacts use regression mode. Their returned scores pass through Equation~\ref{eq:output}. The XGBoost JSON contains eight named features and a \code{binary:logistic} objective, but the application uses the classifier's \code{predict} method, which returns labels. The integrated ensemble is an outer stacking classifier with four inner stacking classifiers corresponding to Random Forest, Gradient Boosting, LinearSVC, and XGBoost. Each inner stack contains estimators named for accuracy, recall, and precision; the supplied configuration uses Logistic Regression as the four inner and outer final estimators. These structures are retained from the supplied artifacts rather than learned by the web service.

Method descriptions must be associated with the actual integrated artifact. In particular, the interface name ``Pruned RuleFit'' is retained without asserting a four-rule-only deployed predictor: the local artifact and the concise method description have not been fully reconciled. Appendix~\ref{app:provenance} records the relevant artifact-level facts. This is an integration-documentation boundary, not a conclusion about the validity of the original pruning research.

\subsection{Field meanings, coding, and validation}
The field order is authoritative for positional construction, while clinical interpretation also depends on definitions and coding. RuleFit includes case-sensitive \code{Hematocrit}; ensemble names include long punctuated keys and use age, gender, and race in addition to laboratory and complication fields. The frontend preserves exact names as keys while mapping many of them to readable labels.

The complication options encode Yes=1 and No=0. The gender selector uses Female=0, Male=1, Unknown=2. Race is represented by nine displayed categories with codes 0--8, listed in Appendix~\ref{app:fields}. These are implemented interface codes; their correspondence to training-data codebooks and the meaning of the source gender field require confirmation. Likewise, the displayed units are entry guidance rather than an implemented unit-conversion layer.

Numerical controls are required, with no explicit \code{min}, \code{max}, or \code{step}. Suggested ranges and age's positive-integer instruction appear in placeholders. The service checks model membership and list type but delegates most value and shape handling to numerical conversion and the estimator. Section~\ref{sec:evaluation} reports the resulting browser and API behavior. A fuller integration contract would couple feature order with units, categorical definitions, allowed values, and missingness semantics.

\subsection{Method context and prediction interpretation}
The application provides method summaries, original-research links, and researcher information beside the form. These features support navigation from an input interface to its research context. RuleFit's rule structure and the elaborative model's feature-selection rationale remain properties of the underlying model research. The current result display presents a binary classification with method context; it does not render patient-specific rules, feature contributions, probabilities, or uncertainty intervals. The helper's common threshold therefore establishes a display contract, not a shared calibrated probability scale. High/Low Risk text is a label for that classification, rather than an individual explanation or a clinical exclusion rule.

\section{System Functional Evaluation}\label{sec:evaluation}
\subsection{Evaluation design}
The completed evaluation was performed on September 8, 2026. It combines three evidence layers: source-level verification of component contracts, real-model execution through Flask's test client, and actual browser interaction with the shared React component using mocked transport. The four-model application snapshot and its supplied artifacts define the evaluated system.

Backend execution used unchanged copies of the application files and artifacts in an isolated Python 3.12.14 environment. The original environment's Python 3.13 startup did not complete the initial probe. An import-only PyTorch shim supported the helper's unused unconditional import; PyTorch loading was unavailable. The test harness also restricted deserialized globals and blocked network operations and subprocesses. These are test-environment adaptations, not application features. Main package versions, differences and diagnostic scope appear in Appendix~\ref{app:environment}.

For each model, three synthetic vectors were used: a hand-constructed numerical baseline, a 5\% perturbation of its non-age numerical measurements with categorical codes retained, and an all-zero vector. Exact baselines and construction rules appear in Appendix~\ref{app:fixtures}. They are software fixtures without patient identities or DR labels. The twelve comparisons check invocation paths and application thresholding. For the nine non-composite cases, both raw paths call the same estimator's \code{predict(x)}; their agreement is not an independent implementation comparison. The three ensemble cases compare explicit \code{model.predict(scaler.transform(x))} with the composite wrapper, then check the helper's thresholded class. This design provides bounded adapter and threshold evidence rather than cross-version equivalence or diagnostic validation.

Flask's test client made 62 requests without a running HTTP server. Cases covered registry and feature retrieval, normal vectors, wrong lengths, type and numerical boundaries, invalid envelopes, and ancillary routes. Browser testing mounted the original form component in an isolated React 19.0.0/Axios harness and used Chrome 152.0.7977.77. Feature responses supplied the verified ordered lists; prediction responses were controlled classes 0 and 1. Distinct values at numerical positions made payload ordering observable. Twenty-four model-related scenarios covered blank input, ordered payloads, result mapping, decimal focus/blur and keyboard behavior, and values outside displayed hints.

\subsection{Model loading, input order, and response behavior}
All four artifacts loaded under the documented adaptations and exposed the expected 14, 6, 8, and 25 fields. The scaler was excluded from the standalone model list and paired with the ensemble. All 12 limited-vector path and threshold checks agreed. The baseline, perturbation, and zero-vector outputs are provided in Appendix~\ref{app:fixtures}; these results describe invocation behavior, not clinical accuracy.

The 62 requests comprise 48 model-dependent prediction cases, four ordered-feature requests, five registry/unknown-feature/ancillary GETs, and five invalid-envelope POSTs. The model-dependent outcomes are shown in Appendix~\ref{app:boundary}. Wrong-length, empty-list, and nonnumeric-value inputs returned 500 for all models, whereas numeric strings, nested one-row lists, negative values, and an out-of-domain neuropathy code were accepted. Every successful prediction response contained an integer 0 or 1. HTTP 200 indicates that the software produced a response, not that an input is medically reasonable.

Unknown models and invalid feature-envelope types returned 400 through the prediction route. Feature retrieval for an unknown model returned 200 with an empty list. A separate missing-scaler check returned an unwrapped stacking classifier without the expected \code{feature_names} attribute. This confirms the loader's fallback behavior and motivates an explicit artifact--preprocessor pairing contract.

\subsection{Browser behavior and representative interface}
The component rendered and submitted all 53 field positions across the four model contracts in their returned order. Required blank forms produced no prediction request. Controlled class 0 and 1 responses produced the corresponding Low/High Risk text on all four forms. This verifies the browser component and JSON construction against known transport responses; it does not establish complete browser-to-model end-to-end operation.

Decimal behavior depended on interaction state. After a prior integer value attribute of 10, entering 1.7 in the focused first numerical field produced \code{stepMismatch=true}; pressing Enter sent no prediction request. After blur, React synchronized the value attribute to 1.7, the control became valid, and button submission transmitted 1.7. An uncontrolled input retained the mismatch, consistent with HTML's default numerical step of 1 \citep{whatwg}. A first value of $-999$ was also submitted on every form. These observations distinguish browser-native required/step behavior from enforcement of the displayed medical hints.

\begin{figure}[p]
\centering\includegraphics[width=\linewidth,height=.90\textheight,keepaspectratio]{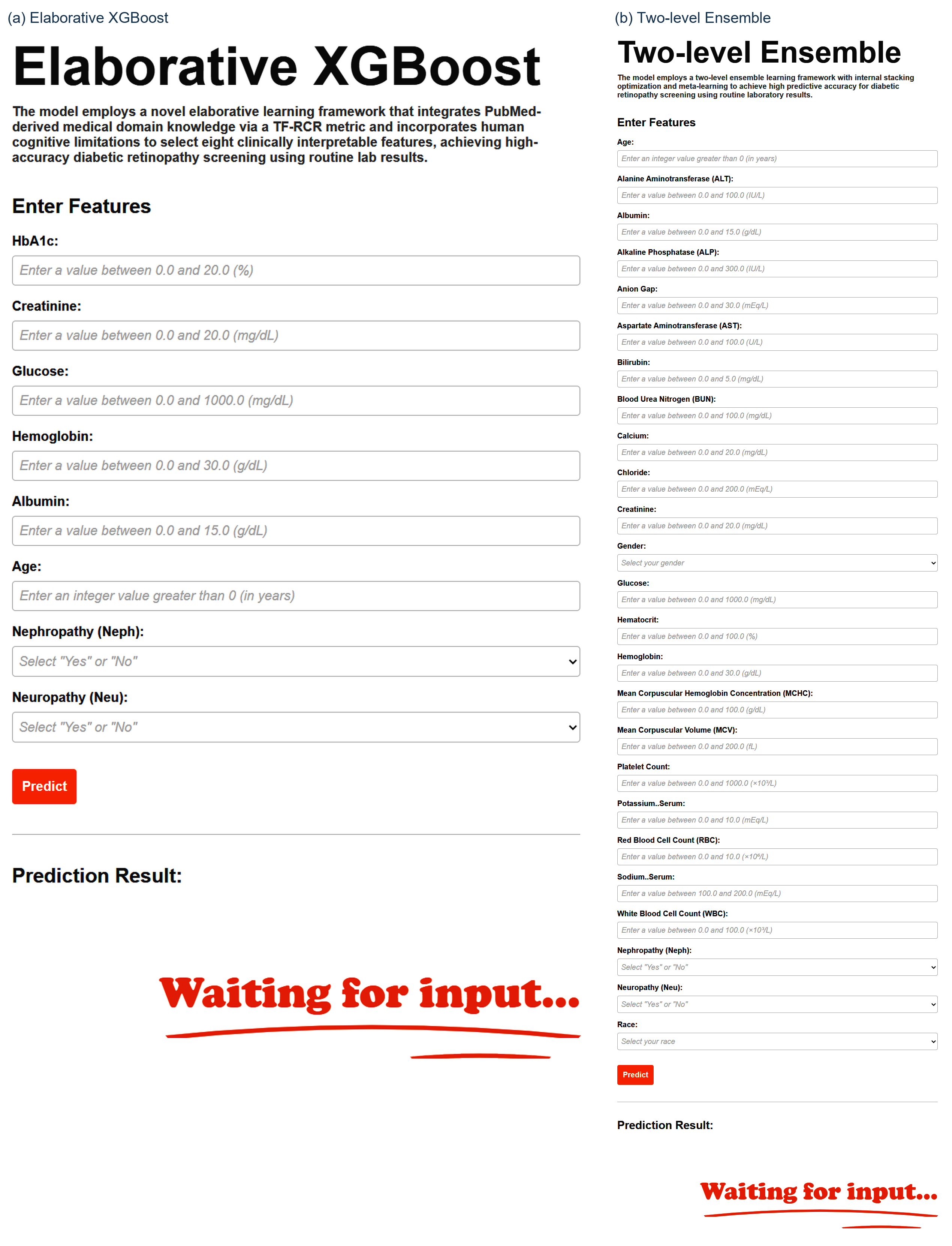}
\caption{Representative pages from the complete four-model DR-LabStack system: (a) Elaborative XGBoost with all 8 input fields (left); (b) Two-level Ensemble with all 25 input fields (right). Both show the complete prediction operation column in the waiting-for-input state.}
\label{fig:interface}
\end{figure}

Figure~\ref{fig:interface} presents two representative pages from the complete four-model system. The left Elaborative XGBoost panel includes all eight input fields, and the right Two-level Ensemble panel includes all 25. Both retain the model title, complete input form, Predict button, Prediction Result heading, and waiting-for-input state. These supplied screenshots illustrate the shared interaction structure and are separate from the executed tests.

\section{Discussion}
\subsection{Separating numerical identity from presentation}
The shared form connects two representations of an input: its exact model key and its displayed clinical label. Using the returned sequence for both rendering and payload construction makes positional correspondence explicit across four different input sizes. Keeping the label map separate permits readable names without rewriting the numerical contract. The browser results support this ordering behavior for the implemented fields. Structural reuse is a design property here; development time, clinical input-error reductions, and usability improvements were not measured.

Feature order is one part of the contract. The same correctly positioned value can have a wrong unit, an unsupported categorical code, or an ambiguous clinical definition. A transferable next step is a typed feature schema that binds those meanings to exact keys, then drives both presentation and service-side validation. Such a schema would also distinguish a required value from an acceptable missing-value representation. The current separation between artifact metadata and frontend hints makes these responsibilities visible. This is consistent with Sculley et al.'s analysis of input dependencies: a change in an input's meaning can affect the consuming model even when the software connection remains intact \citep{sculley2015}.

\subsection{Adapting artifacts while retaining preprocessing}
A common prediction interface allows the service to accommodate JSON and Python-serialized models, with an additional composite path for external preprocessing. The ensemble checks directly exercise that composite path. Maintaining this behavior over time requires more than adding a file: artifact and scaler identity, metadata origin, compatible dependencies, and page configuration must remain aligned. A versioned integration manifest could bind these elements without treating model training as part of online serving.

Source and researcher links are useful alongside this technical contract because they preserve attribution and offer method context. Their association with a specific artifact should be maintained when an integration changes. This responsibility is distinct from reassessing the original researchers' algorithmic claims or assigning a published experiment's performance to an unmatched binary.

\subsection{Uniform responses and meaningful outputs}
A shared response field and classification display simplify the interface across heterogeneous estimators. The underlying \code{predict} outputs nevertheless differ: the RuleFit artifacts provide regression scores, while the classifier interfaces provide labels. Equation~\ref{eq:output} standardizes the final class, not the evidential meaning of that class. Future probability or explanation displays would require model-specific output definitions and validation before being added to the common response schema.

This distinction matters for a clinician-facing application. Labels, method context, and the actual prediction should have a clearly stated role in the workflow. A low classification cannot establish absence of DR. Age, complication codes, gender, and race also require context about ascertainment and population; their inclusion is not evidence of a causal mechanism. Usability evaluation should therefore assess interpretation as well as successful data entry. The capability/reliability distinction and clinical onboarding research suggest concrete future tasks: asking users to identify required inputs, explain the binary result, and describe the intended limits of use \citep{amershi2019hai,cai2019}. These are proposed evaluation tasks; successful performance on them has not been established for this system.

\section{Limitations and Future Work}
The functional evidence covers one four-model snapshot and a limited set of software inputs. Isolated execution differed from the original environment in Python and several dependencies, with explicit test-only adaptations. Browser tests used mocked transport in one Chrome version, and Flask test-client calls did not exercise a live full-stack HTTP deployment. The study therefore establishes the tested component and service behaviors within those boundaries.

This work evaluates the integration and serving behavior of externally developed pretrained models. Model development and training were outside its scope, and no independent diagnostic validation was performed. Available handoff materials did not establish a complete mapping from every integrated artifact to a specific published experiment; consequently, performance estimates from the original studies are not attributed to the deployed artifacts. This limitation concerns the evidence available for system integration, rather than whether the original researchers performed particular procedures.

Further engineering work should prioritize a unified feature schema, server-side validation of input shape, type and values, explicit binding of models to preprocessing objects, a reproducible dependency environment, and clearer error responses. The observed missing-scaler fallback and browser decimal behavior provide concrete cases for these changes. Serialization compatibility requires attention because successful loading under a newer library does not establish agreement with the training runtime \citep{sklearnpersistence}. Real frontend--backend integration tests and cross-browser checks should follow within a fixed application version.

A further input pathway could support clinician-reviewed extraction of information from clinical narratives. AutoPCR combines entity extraction, candidate concept retrieval, and ontology-based entity linking to recognize phenotype concepts \citep{tao2026autopcr}. Such recognition could supply candidate phenotype fields for review, while mapping to model-specific features would still require handling negation, time context, numerical values, units, and missingness. This is a proposed extension to the current manual-entry workflow; no text-extraction component was implemented or evaluated here.

Future model adapters could also accommodate paired retinal images and clinical metadata. RiVaT-Fuse provides an example of reliability-aware image--metadata fusion \citep{xu2026rivat}, while MobiFusion-DRM combines these inputs for DR severity, macular edema, and referable-status prediction \citep{xu2026mobifusion}. Serving such models would require image upload and preprocessing, explicit image--metadata pairing, modality-availability checks, and task-specific response fields and displays. For multi-output models, the interface should preserve the distinction between ordinal grades and binary findings, with any probabilities or thresholds defined per task. These models are prospective integration examples; neither was incorporated into the four-model system evaluated here.

Beyond input validation, future deployments should examine whether incoming patient populations differ from the data supporting the integrated models. Detectron formulates a hypothesis test for harmful covariate shift using disagreement among trained classifiers \citep{ginsberg2023detectron}, offering a methodological reference for this monitoring problem. The monitoring target should be specified: shifts within an existing label space differ from detecting samples outside the training categories. For the latter task, Lu et al. distinguish out-of-distribution detection approaches by the ability to modify or retrain the prediction model \citep{lu2025out}. This access requirement matters when selecting monitoring methods for externally supplied artifacts. Suitable reference data and separate evaluations would be needed; neither detection task is assessed by the current synthetic-input checks.

The current implementation is a research web system: its hardcoded loopback HTTP endpoints, permissive CORS, debug-mode startup and limited operational controls require deployment-specific engineering. Deployment performance and clinical-workflow use have not been measured. Future work should evaluate task completion, input errors, result comprehension and satisfaction with intended clinical users under a documented protocol. Any subsequent diagnostic evaluation should be conducted with the original model researchers, using an authorized cohort and a defined target and validation design. Usability and clinical studies should follow appropriate ethics and stage-specific reporting procedures \citep{collins2024,vasey2022}.

\section{Conclusion}
We designed and implemented DR-LabStack as a common web workflow for four externally developed DR prediction models. The application connects model-derived feature order to shared React controls, adapts heterogeneous artifacts and the ensemble scaler through Flask, and presents a common classification response with research context. Real-model service checks and separate browser-component tests establish the reported loading, ordering, invocation and display behavior under documented conditions. The system provides a concrete foundation for further integration engineering and clinician-facing evaluation, with usability, deployment performance and clinical effectiveness remaining to be assessed.

\section*{Data and Code Availability}
The project repository is available at \href{https://github.com/TerrificXu/diabetic-retinopathy-web}{TerrificXu/diabetic-retinopathy-web}. Public availability and permission to reuse individual components are separate matters; a uniform license over code, model artifacts and third-party assets is not asserted here. The functional evaluation used synthetic software vectors rather than patient records. This manuscript and its appendices specify the principal input contracts, fixtures, environment and outcomes. No training dataset or complete training-pipeline release is claimed, and no patient data or model binaries are included in the manuscript source package.

\section*{Ethics and Declarations}
The software functional evaluation used synthetic inputs, recruited no participants, and processed no patient records. Training and clinical validation of the externally developed models were outside this study's scope. The present software evaluation makes no claim about ethics approval for those separate studies.

\paragraph{System development.} Yingfan Xu was responsible for web-system design, implementation, integration, and software verification. The underlying prediction models were developed in separate research, as attributed in this manuscript.

\paragraph{AI assistance.} OpenAI Codex assisted with manuscript drafting and editing, literature verification, evaluation-tool preparation and execution, figure preparation, and document compilation and checking. The reported functional results are the recorded software-test outcomes. The authors are responsible for the submitted manuscript.

\bibliographystyle{plainnat}
\begingroup
\interlinepenalty=10000
\bibliography{references}
\endgroup
\appendix
\input{appendix_contracts}
\end{document}

%% file: appendix_contracts.tex
\clearpage
\section{Ordered Input Fields and Display Definitions}\label{app:fields}
Exact keys retain their original spelling and order. The accompanying labels, units and codes describe the implemented interface; compatibility with source-data definitions requires confirmation. Numerical inputs are required, with suggested ranges in placeholders and no explicit min/max/step attributes. Complication selectors use Yes=1 and No=0.

\noindent\begin{minipage}{\linewidth}
\subsection*{RuleFit (14)}
\begingroup\fontsize{9}{11}\selectfont
\begin{tabular}{@{}rP{.488\linewidth}P{.43\linewidth}@{}}
\toprule No. & Exact model key & Display meaning / unit or encoding\\\midrule
1 & \code{hba1c} & HbA1c; \%\\
2 & \code{creatinine} & Creatinine; mg/dL\\
3 & \code{neu} & Neuropathy (Neu); Yes=1; No=0\\
4 & \code{Hematocrit} & Hematocrit; \%\\
5 & \code{bun} & Blood Urea Nitrogen (BUN); mg/dL\\
6 & \code{neph} & Nephropathy (Neph); Yes=1; No=0\\
7 & \code{albumin} & Albumin; g/dL\\
8 & \code{calcium} & Calcium; mg/dL\\
9 & \code{sodium} & Sodium; mEq/L\\
10 & \code{anion_gap} & Anion Gap; mEq/L\\
11 & \code{alt} & Alanine Aminotransferase (ALT); U/L\\
12 & \code{bilirubin} & Bilirubin; mg/dL\\
13 & \code{chloride} & Chloride; mEq/L\\
14 & \code{potassium} & Potassium; mEq/L\\
\bottomrule\end{tabular}\endgroup
\end{minipage}\par\medskip

\noindent\begin{minipage}{\linewidth}
\subsection*{Pruned RuleFit (6)}
\begingroup\fontsize{9}{11}\selectfont
\begin{tabular}{@{}rP{.488\linewidth}P{.43\linewidth}@{}}
\toprule No. & Exact model key & Display meaning / unit or encoding\\\midrule
1 & \code{creatinine} & Creatinine; mg/dL\\
2 & \code{neu} & Neuropathy (Neu); Yes=1; No=0\\
3 & \code{hba1c} & HbA1c; \%\\
4 & \code{bun} & Blood Urea Nitrogen (BUN); mg/dL\\
5 & \code{neph} & Nephropathy (Neph); Yes=1; No=0\\
6 & \code{anion_gap} & Anion Gap; mEq/L\\
\bottomrule\end{tabular}\endgroup
\end{minipage}\par\medskip

\noindent\begin{minipage}{\linewidth}
\subsection*{Elaborative XGBoost (8)}
\begingroup\fontsize{9}{11}\selectfont
\begin{tabular}{@{}rP{.488\linewidth}P{.43\linewidth}@{}}
\toprule No. & Exact model key & Display meaning / unit or encoding\\\midrule
1 & \code{hba1c} & HbA1c; \%\\
2 & \code{creatinine} & Creatinine; mg/dL\\
3 & \code{glucose} & Glucose; mg/dL\\
4 & \code{hemoglobin} & Hemoglobin; g/dL\\
5 & \code{albumin} & Albumin; g/dL\\
6 & \code{age} & Age; in years\\
7 & \code{neph} & Nephropathy (Neph); Yes=1; No=0\\
8 & \code{neu} & Neuropathy (Neu); Yes=1; No=0\\
\bottomrule\end{tabular}\endgroup
\end{minipage}\par\medskip

\clearpage

\subsection*{Two-level Ensemble (25)}
\begingroup\fontsize{9}{11}\selectfont
\renewcommand{\theHtable}{contracts.\arabic{table}}
\begin{longtable}{@{}rP{.488\linewidth}P{.43\linewidth}@{}}
\toprule No. & Exact model key & Display meaning / unit or encoding\\\midrule\endfirsthead
\toprule No. & Exact model key & Display meaning / unit or encoding\\\midrule\endhead
1 & \code{AGE_IN_YEARS} & Age; in years\\
2 & \code{Alanine.Aminotransferase...SGPT} & Alanine Aminotransferase (ALT); IU/L\\
3 & \code{Albumin..Serum} & Albumin; g/dL\\
4 & \code{Alkaline.Phosphatase..Serum} & Alkaline Phosphatase (ALP); IU/L\\
5 & \code{Anion.Gap..Blood} & Anion Gap; mEq/L\\
6 & \code{Aspartate.Aminotransferase} & Aspartate Aminotransferase (AST); U/L\\
7 & \code{Bilirubin.Total.Serum.or.Plasma.Mass.Volume} & Bilirubin; mg/dL\\
8 & \code{Blood.Urea.Nitrogen} & Blood Urea Nitrogen (BUN); mg/dL\\
9 & \code{Calcium..Serum} & Calcium; mg/dL\\
10 & \code{Chloride..Serum} & Chloride; mEq/L\\
11 & \code{Creatinine..Serum.Quantitative} & Creatinine; mg/dL\\
12 & \code{GENDER} & Gender; codes defined below\\
13 & \code{Glucose..Serum.Plasma.Quantitative} & Glucose; mg/dL\\
14 & \code{Hematocrit} & Hematocrit; \%\\
15 & \code{Hemoglobin} & Hemoglobin; g/dL\\
16 & \code{Mean.Corpuscular.Hemoglobin.Concentration} & Mean Corpuscular Hemoglobin Concentration (MCHC); g/dL\\
17 & \code{Mean.Corpuscular.Volume} & Mean Corpuscular Volume (MCV); fL\\
18 & \code{Platelet.Count} & Platelet Count; $\times$10$^3$/L\\
19 & \code{Potassium..Serum} & Potassium..Serum; mEq/L\\
20 & \code{Red.Blood.Cell.Count} & Red Blood Cell Count (RBC); $\times$10$^6$/L\\
21 & \code{Sodium..Serum} & Sodium..Serum; mEq/L\\
22 & \code{White.Blood.Cell.Count} & White Blood Cell Count (WBC); $\times$10$^3$/L\\
23 & \code{nephropathy} & Nephropathy (Neph); Yes=1; No=0\\
24 & \code{neuropathy} & Neuropathy (Neu); Yes=1; No=0\\
25 & \code{race} & Race; codes defined below\\
\bottomrule\end{longtable}\addtocounter{table}{-1}\endgroup

Gender is encoded Female=0, Male=1, Unknown=2. Race codes are African American=0, Asian=1, Biracial=2, Caucasian=3, Hispanic=4, Eastern Indian=5, Native American=6, Other=7, and Pacific Islander=8. Source-data meanings and coding require confirmation. The platelet/WBC hints show $\times 10^3$/L and the RBC hint $\times 10^6$/L as implemented; these are reproduced rather than medically corrected. Some serum keys retain their original name as a display fallback.

\section{Functional Evaluation Details}
\subsection{Environment and adaptation scope}\label{app:environment}
The September 8, 2026 run used Windows 11 build 26200 and Python 3.12.14. Package versions were NumPy 2.3.1, SciPy 1.18.1, scikit-learn 1.7.0, pandas 2.3.0, XGBoost 3.0.2, joblib 1.5.1, RuleFit 0.3.1, Flask 3.1.1, Flask-CORS 6.0.1, Flask-WTF 1.2.2, WTForms 3.2.2, and Werkzeug 3.1.8. The inspected original environment used Python 3.13, SciPy 1.16.0, WTForms 3.2.1 and Werkzeug 3.1.3. The current artifacts include serialized scikit-learn 1.6.0 (RuleFit) and 1.6.1 (pruned model and ensemble/scaler) metadata; the XGBoost JSON records version 2.0.3.

Copied application and artifact bytes were unchanged. The harness introduced an import-only torch shim, disabled PyTorch artifact loading, restricted deserialization globals, and prevented network/subprocess calls. The shim was removed from global module discovery after helper import. Initial harness failures due to a narrow allowlist and shim handling were corrected in testing code only; they were not application repairs or counted as successful original-environment runs. The successful evaluation applies to this adapted environment. Browser tests used Node 24.19, React/ReactDOM 19.0.0, Axios 1.7.9 and Chrome 152.0.7977.77, with the shared component transpiled into an isolated root and feature/prediction transport intercepted.

\subsection{Synthetic fixtures and comparison definition}\label{app:fixtures}
The following baseline vectors follow the field order in Appendix~\ref{app:fields}. A second vector multiplies numerical measurements by 1.05, retaining age, gender, race and complication values; a third sets all positions to zero. These inputs contain no outcome label, and zero or perturbed values may be medically implausible. For non-composite models, direct and raw paths both call the same estimator. For the ensemble, the direct path explicitly scales then predicts, while the raw path calls CompositeModel. In every case, the helper's returned class is compared with a strict $>0.5$ threshold applied to the first direct output.

\noindent\textbf{RuleFit baseline:}\par
{\small [7, 1, 0, 40, 20, 0, 4, 9, 140, 12, 20, 1, 100, 4]}\par

\noindent\textbf{Pruned RuleFit baseline:}\par
{\small [1, 0, 7, 20, 0, 12]}\par

\noindent\textbf{Elaborative XGBoost baseline:}\par
{\small [7, 1, 120, 14, 4, 50, 0, 0]}\par

\noindent\textbf{Two-level Ensemble baseline:}\par
{\small [50, 20, 4, 80, 12, 20, 1, 20, 9, 100, 1, 0, 120, 40, 14, 34, 90, 250, 4, 5, 140, 7, 0, 0, 3]}\par

\begin{table}[ht]
\centering\small
\caption{Retained fixture results, shown as first raw output / helper class. Decimal scores are rounded here to six places; equality checks used unrounded arrays. Each row represents three software inputs, not patients.}
\begin{tabular}{llll}
\toprule Model & Baseline & Perturbed & All-zero\\\midrule
RuleFit & 0.369084 / 0 & 0.378369 / 0 & 0.080717 / 0\\
Pruned RuleFit & 0.323701 / 0 & 0.390906 / 0 & 0.075631 / 0\\
Elaborative XGBoost & 0 / 0 & 1 / 1 & 0 / 0\\
Two-level Ensemble & 0 / 0 & 0 / 0 & 1 / 1\\
\bottomrule\end{tabular}\end{table}

\subsection{Model-dependent API outcomes}\label{app:boundary}
The 48 prediction requests comprise the twelve conditions below for each of four models. Each test changes the baseline vector as named: nonnumeric first value is ``bad'', numeric strings apply to every position, nested input is a one-row list, negative first value is $-999$, and neuropathy code is 9. Null replaces the first value. NaN/Infinity stress requests use the permissive Python JSON encoder and are nonstandard JSON; they are distinct from ordinary browser JSON. Outcomes below preserve failures and acceptance behavior as observed.
\begin{table}[ht]
\centering\small
\caption{HTTP status of model-dependent test-client cases. RF: RuleFit; PRF: Pruned RuleFit; ENS: Two-level Ensemble. HTTP 200 is software acceptance, not medical validity.}
\begin{tabular}{lrrrr}
\toprule Input condition & RF & PRF & XGB & ENS\\\midrule
Baseline numerical vector & 200 & 200 & 200 & 200\\
Last field removed & 500 & 500 & 500 & 500\\
One extra value (1) & 500 & 500 & 500 & 500\\
Nonnumeric first value & 500 & 500 & 500 & 500\\
Empty feature list & 500 & 500 & 500 & 500\\
Numeric strings & 200 & 200 & 200 & 200\\
Nested one-row list & 200 & 200 & 200 & 200\\
Negative first value & 200 & 200 & 200 & 200\\
Neuropathy code 9 & 200 & 200 & 200 & 200\\
Null first value & 500 & 500 & 200 & 500\\
NaN first value (nonstandard JSON) & 500 & 500 & 200 & 500\\
Infinity first value (nonstandard JSON) & 200 & 200 & 200 & 500\\
\bottomrule\end{tabular}\end{table}

The other 14 requests were four metadata-order checks; GETs to models, unknown-model features, the legacy root, about, and its stylesheet; and five prediction envelopes: empty object, unknown model with a list, known model with a string, missing features, and null features. The latter five returned 400. The 24 browser scenarios comprise six per model: required-empty submission, ordered payload/class 0, class 1 display, focused-decimal Enter, blurred-decimal button submission, and negative-value submission. Numerical positions used distinct values starting at 10; selectors used valid option codes. Two additional non-model probes characterized JavaScript conversion and an uncontrolled number input. They are outside the 24 model-scenario count.

\subsection{Artifact-to-description scope}\label{app:provenance}
The current RuleFit object stores 1,443 rule terms and 14 linear terms, of which 110 and 10 respectively have nonzero coefficients. The Pruned RuleFit object stores 1,452 rule terms and 6 linear terms, with 91 and 4 nonzero coefficients. These are stored term counts, not per-patient active-rule counts. They qualify the local four-rule interface description without establishing a conclusion about the underlying pruning study. The integrated ensemble has Logistic Regression final estimators; the source study reports multiple configurations. Artifact-to-experiment correspondence remains incomplete, so its principal Random Forest configuration's performance is not assigned to this artifact.

\subsection{Ancillary application paths}\label{app:ancillary}
Flask-WTF templates coexist with the React interface. The legacy root returned 200, its about route 500 because the referenced template was absent, and its stylesheet 404. The JSON model-list endpoint does not render the separate models.html template. Standalone ModelSelection and PredictionResult components are outside the active prediction route graph. The contact component sends name/email/subject/message/body while the backend expects recipient/user\_email/subject/body. This static contract mismatch was not tested by sending email. These supporting-path findings are separate from the four-model inference results.